\documentclass[11pt]{article}

\usepackage{acl}

\usepackage{times}
\usepackage{latexsym}
\usepackage{makecell}
\usepackage{multirow}
\usepackage{graphicx}
\usepackage{hyperref}
\usepackage{float}
\usepackage{booktabs}
\usepackage{enumitem}
\usepackage{tabularx} 
\usepackage{array}
\newcolumntype{Y}{>{\centering\arraybackslash}X}
\usepackage{siunitx}    
\usepackage{booktabs,tabularx,array,makecell}
\newcolumntype{Y}{>{\centering\arraybackslash}X}           
\newcolumntype{L}[1]{>{\raggedright\arraybackslash}p{#1}} 
\usepackage[T1]{fontenc}

\usepackage[utf8]{inputenc}

\usepackage{microtype}

\usepackage{inconsolata}

\usepackage{graphicx}

\usepackage[most]{tcolorbox}
\usepackage{xcolor}

\definecolor{rqblue}{RGB}{100,160,220}
\definecolor{rqbluebg}{RGB}{240,248,255}
\definecolor{rqgreen}{RGB}{120,190,120}
\definecolor{rqgreenbg}{RGB}{242,250,242}
\definecolor{rqorange}{RGB}{240,170,100}
\definecolor{rqorangebg}{RGB}{255,248,240}
\definecolor{rqpurple}{RGB}{170,140,200}
\definecolor{rqpurplebg}{RGB}{248,245,252}
\definecolor{rqred}{RGB}{220,100,100}
\definecolor{rqredbg}{RGB}{255,245,245}

\newcommand{\recommendation}[4][]{%
  \begin{tcolorbox}[
    colframe=rq#3,
    colback=rq#3bg,
    coltitle=white,
    fonttitle=\bfseries,
    fontupper=\color{black},
    boxrule=0.8pt,
    arc=3pt,
    left=5pt,right=5pt,top=5pt,bottom=5pt,
    title={\if\relax\detokenize{#1}\relax Recommendation~#2\else #1~#2\fi}
  ]
  #4
  \end{tcolorbox}
}

\usepackage{todonotes}

\usepackage{xcolor} 

\newif\ifcomment
\commenttrue   

\ifcomment
  \newcommand{\haonan}[1]{\textcolor[rgb]{0.6,0,0.2}{haonan: #1}}
  \newcommand{\jiefu}[1]{\textcolor{red}{Zi Ang: #1}}
  \newcommand{\junfeng}[1]{\textcolor[blue]{0.3,0.7,0.2}{junfeng: #1}}
  
\else
  \newcommand{\haonan}[1]{}
  \newcommand{\jiefu}[1]{}
  \newcommand{\junfeng}[1]{}
\fi

\title{ByteSized32Refactored: Towards an Extensible Interactive Text Games Corpus for LLM World Modeling and Evaluation}

\author{
  Haonan Wang\textsuperscript{1}, 
  Junfeng Sun\textsuperscript{2}, 
  Xingdi Yuan\textsuperscript{3},
  Ruoyao Wang\textsuperscript{4},
  Ziang Xiao\textsuperscript{1} \\
  \textsuperscript{1}Johns Hopkins University\quad
  \textsuperscript{2}Liaoning Technical University\quad
  \textsuperscript{3}Microsoft Research Montréal\quad \\ 
  \textsuperscript{4}Central University of Finance and Economics \\
  \texttt{hwang298@jh.edu} \quad \texttt{junf1831@outlook.com} \quad \texttt{eryua@microsoft.com}\\
  \quad \texttt{wangruoyao@cufe.edu.cn}\quad \texttt{ziang.xiao@jhu.edu}
}

\begin{document}
\maketitle
\begin{abstract}
Simulating interactive world models remains a core challenge in Large Language Models(LLMs). In this work, we introduce the \textbf{ByteSized32Refactored}\footnote{ByteSized32Refactored:\url{https://github.com/isle-dev/BYTESIZED32-Refactored}}, a refactored, modular, and extensible implementation of the original ByteSized32\footnote{ByteSized32:\url{https://github.com/cognitiveailab/BYTESIZED32}} corpus to explore the task of text game generation. We further optimize the code structure of each text game and create the \verb|GameBasic.py| foundation library, which centralizes common logic across all 32 games by abstracting 7 base classes (\texttt{GameObject}, etc.) into reusable modules, thereby reducing from \textbf{20k} to \textbf{10k} total lines of Python code compared to the original Bytesized32. Our refactored implementation enables extendability - with our centralized design, \textbf{ByteSized32Refactored} can be more efficiently extended to include text games of new scenarios and specifications by reusing the shared logic and functionalities. Extensive experiments with GPT-4o demonstrate a mix of performance - with \textbf{Bytesized32Refactored}, the generated text games for unseen scenarios showcase quality improvements on two of the four evaluation dimensions while decreases on the other two, indicating that the hierarchical structure of the refactored code presents new challenges for LLMs.
Overall, we highlight that our extensible code structure, centered on the foundation library and the modular optimization, not only facilitates LLM adaptation to environment specifications but also establishes a scalable environment that supports future extensions. 
\end{abstract}

\section{Introduction}
Human intelligence and problem-solving are characterized by the ability to understand and interact with
\begin{figure*}[htbp]
    \centering \includegraphics[width=0.9\linewidth,keepaspectratio,,trim=62 15 80 40,clip]{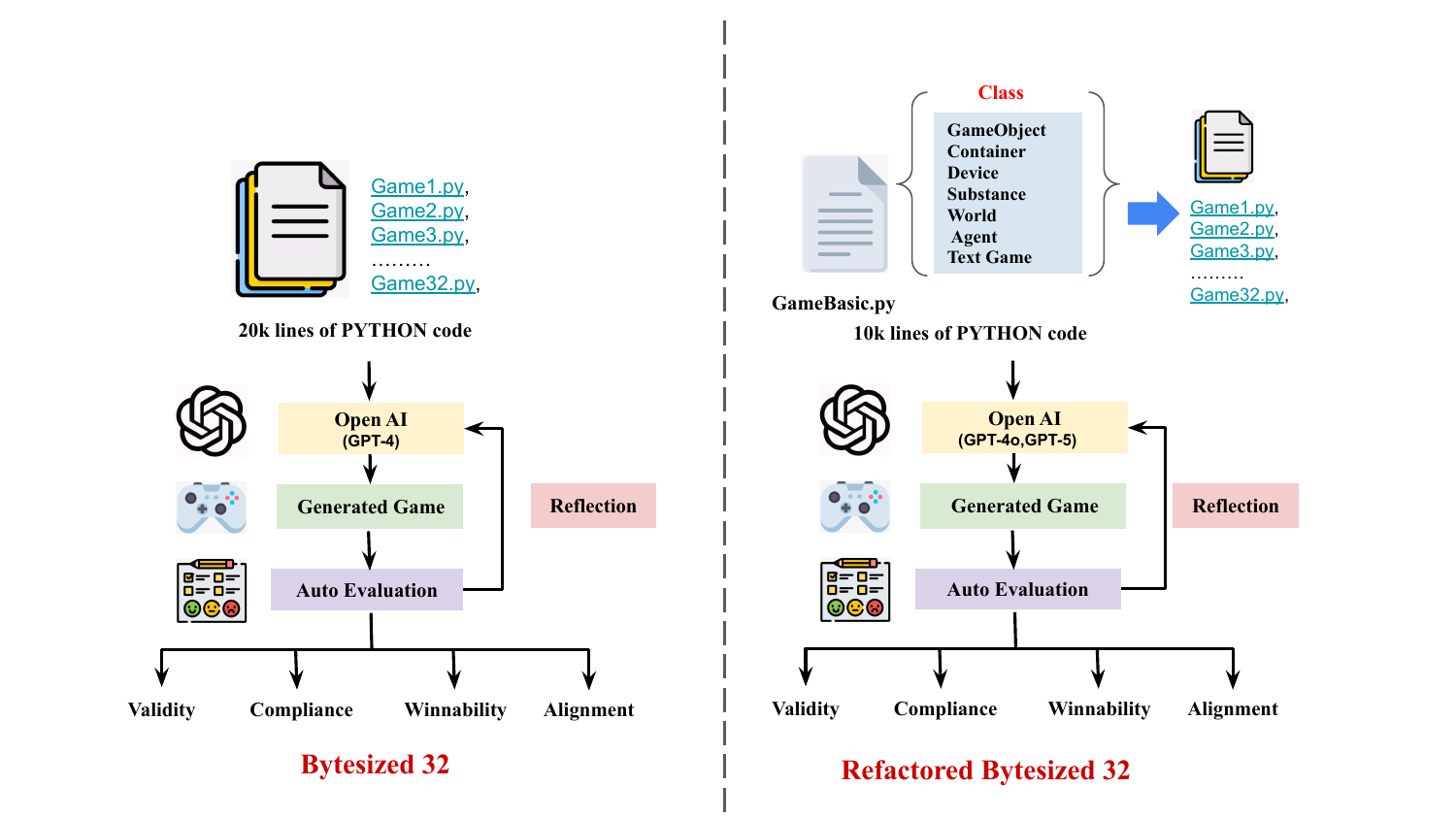}
    \caption{An overview of text game generation and evaluation process between the \textbf{ByteSized32Refactored} and ByteSized32. GPT-4o and GPT-5 generate games using in-context learning with a prompt consisting of (1) a single template example game, and (2) the task specification for the target game to generate. Generated games are then self-reflected by providing the models with error outputs from a PYTHON interpreter that detects syntactic and API issues. Each generated game is subsequently evaluated by an automated evaluation for technical validity, specification compliance, physical reality alignment, and winnability.}
\end{figure*}
structured representations of the world, which support reasoning, planning, and generalization across complex environmental tasks \cite{gignac2024defining,barsalou1999perceptual}. Large language models (LLMs) have demonstrated impressive capabilities in natural language processing and zero-shot performance across task \cite{kojima2022large}. While these abilities indicate a nascent capacity for LLMs to possess the foundations of world modeling, their reasoning remains largely textual and lacking explicit formalization. 
Using LLMs to play text games primarily assesses their reasoning under constrained specifications, rather than whether they maintain consistent and structured models of the world \cite{Jansen2022TextWorldExpressST, aidungeon, Hausknecht2019InteractiveFG}. To move beyond surface-level evaluation, we argue that generating interactive world simulators in text games provides a more suitable medium for examining world models in LLMs. It offers a structured, interpretable, and testable representation of environments as they formalize objects, states, constraints, and goals with natural language description \cite{wang2023bytesized32corpuschallengetask, wang2022scienceworldagentsmarter5th}.
ByteSized32\cite{wang2023bytesized32corpuschallengetask, wang2022scienceworldagentsmarter5th} 
marked an advanced step in formalizing world modeling, which focuses on the challenge of building text-based environments, where players interact with the world through natural language descriptions and commands. Such text-based environments enable agents to effectively operate within interactive text games to accomplish goals in complex scenarios through natural language input. Its highlights make it an ideal testing environment for evaluating the ability of an AI system in understanding, manipulating and constructing structured representations of the world. However, despite its significance, ByteSized32 also presents notable limitations in extendable code structure design: each game is implemented as a standalone file with logic, which hampers extensibility (adding new actions/objects/rules) and better input formulation for LLMs, i.e. flexible single-shot and multi-shot evaluation within a fixed context window, to generate new text games.
In this work, we present \textbf{ByteSized32Refactored}, a restructured, extensible text game corpus built upon the original  \href{https://github.com/cognitiveailab/BYTESIZED32}{ByteSized32}, designed to evaluate the ability of multiple large language models(LLMs) like   
GPT-4o\footnote{\url{https://openai.com/index/gpt-4o-system-card/}} to construct and test task-specific world environments. Specifically, compared with the original ByteSized32, by reorganizing the foundation library 
\verb|GameBasic.py| that abstracts the 7 base classes (e.g. \textit{GameObject}, \textit{Container}, \textit{World}, \textit{Agent}, and \textit{TextGame}, etc.) and optimizing the code details of each game file. As a result, each game in \textbf{RefactoredBytesized32} now only implements domain-specific objects and task logic, while all shared functionality is reused from \verb|GameBasic.py|, ensuring modularity, extensibility, and reproducibility across the corpus. Moreover, the refactorization enhances the extendability of Bytesized32 to incorporate new text games with novel specifications, improves the utility of existing text games as demonstrations for in-context game generation, and presents a new challenge to LLMs for code generation with hierarchical structures.
\indent The contributions of this work are as follows:
\begin{enumerate}[nosep]
    \item We present \textbf{ByteSized32Refactored}, a modular refactorization of the original ByteSized32 text game corpus (expressed as common-sense text games in Python). In contrast to the original \textit{\textbf{20k}} lines monolithic implementation, our refactored corpus further optimized each game by replacing verbose control flows with dictionary-based dispatch, streamlining string operations, and abstracting redundant logic to comprise \textit{\textbf{10k}} lines (\autoref{sec: eachgamemodular}).
    \item We develop \verb|GameBasic.py| (\autoref{sec: GameBaisc}, \autoref{fig:GameBasic}), a foundational library that provides a shared abstraction layer for all 32 text games, ensuring structural consistency, reducing redundant code, and enabling scalable extensions for future development.
    \item We evaluate GPT-4o on the refactored and original corpus, providing a fair basis for measuring progress over prior results (\autoref{sec: results}).
\end{enumerate}

\section{Related Work}
\subsection{Text Game}
Text games, also known as interactive fiction (IF) environments are simulate interactive worlds in which both observations and actions are expressed entirely in natural language. These text games costs much less than for 2D or 3D games and provides natural language game descriptions of the current state\cite{jansen2021systematic,li2021implicit,nelson2006natural,wang2025design}, which require users or agents to understand and generate natural language commands to interact with the game environment, which have been widely used as a challenging testbed for natural language processing to evaluate multiple capabilities of AI systems\cite{cote2018textworld,wang2023bytesized32corpuschallengetask,wang2022scienceworldagentsmarter5th, Jansen2022TextWorldExpressST,wang2024can,shridhar2020alfworld}. Two dominant approaches have emerged in evaluating AI systems to navigate and complete text world games: reinforcement learning \textbf{(RL)-based agents} modeling the environment as a POMDP \cite{kaelbling1998planning} and large language model \textbf{(LLM)-based agents}, where observations are input to the LLMs and the outputs are executed as actions\cite{cui2025tales}. For \textbf{(RL)-based agents}, prior works have explored various techniques to enhance the learning process and performance for non-LLM-based agents. Subsequent efforts augmented RL agents with structured knowledge representations, such as knowledge graphs, to improve state tracking and filter irrelevant actions\cite{narasimhan2015language,hausknecht2020interactive,ammanabrolu2018playing,yuan2018counting,murugesan2021efficient,ryu2023minimal}. More recent studies have introduced advanced mechanisms, including graph attention layers for more efficient action selection\cite{ammanabrolu2020graph}, role-playing agents that intrinsically reward adherence to personas\cite{peng2023story}, and soft prompts to enable a single agent to adopt multiple personas\cite{cui2023thespian} flexibly. In addition to these purely RL-based methods, recent work has explored hybrid approaches that integrate LLMs with RL agents.\cite{basavatia2024starling}employ LLMs to procedurally generate novel text game environments, thereby enabling RL agents to be tested on generalization across unseen tasks. Similarly, \cite{golchha2024language} utilizes LLMs to provide decision-level guidance to RL agents, improving their reasoning and action selection.

\subsection{LLMs for Code Generation}
Large Language models (LMs) are rapidly being deployed in commercial applications, and several recent works for evaluated the ability of LLMs to generate executable program code\cite{dong2025surveycodegenerationllmbased,huynh2025largelanguagemodelscode}. 
The current standard benchmarks HumanEval\cite{chen2021evaluating}, MBPP\cite{austin2021program}, and APPS\cite{hendrycks2021measuring} are in a longstanding pursuit of synthesizing code from natural language descriptions\cite{yu2018spider,li2022competition,zan2022large}. Similarly, SWE-bench\cite{jimenez2024swebenchlanguagemodelsresolve} has advanced the field by framing software engineering tasks in more realistic contexts, requiring models to navigate large codebases, generate patches, and capture dependency-based relationships across modules. LiveCodeBench\cite{jain2024livecodebench} evaluates models on real-world code changes from open-source projects, while RepoBench\cite{liu2023repobench} focuses on repository-level code completion across multiple files. BigCodeBench\cite{zhuo2024bigcodebench} emphasizes compositional reasoning by requiring function calls across a wide range of libraries, and CoCo-Bench\cite{yin2025coco}extends evaluation to tasks such as code understanding and review. In addition to these benchmarks, several works have explored a variety of extensions have been proposed to broaden its coverage. These include multi-language variants\cite{cassano2022multipl,athiwaratkun2022multi,orlanski2023measuring}, modifications to the edit scope and task granularity\cite{yu2024codereval,du2023classeval}, as well as novel code completion benchmarks\cite{muennighoff2023octopack,liu2023codegeneratedchatgptreally}. Other efforts introduce alternative coding paradigms\cite{yin2022naturallanguagecodegeneration,yang2023intercodestandardizingbenchmarkinginteractive}. However, most of these benchmarks remain limited to short code snippets and fail to capture the complexity of executable programs. They rarely treat code as a fully interactive environment for simulating the world.

\section{Bytesized32Refactored Corpus}
\subsection{Corpus Overview}
\begin{figure*}[t]
    \centering    \includegraphics[width=\linewidth,trim=20 5 10 5,clip]{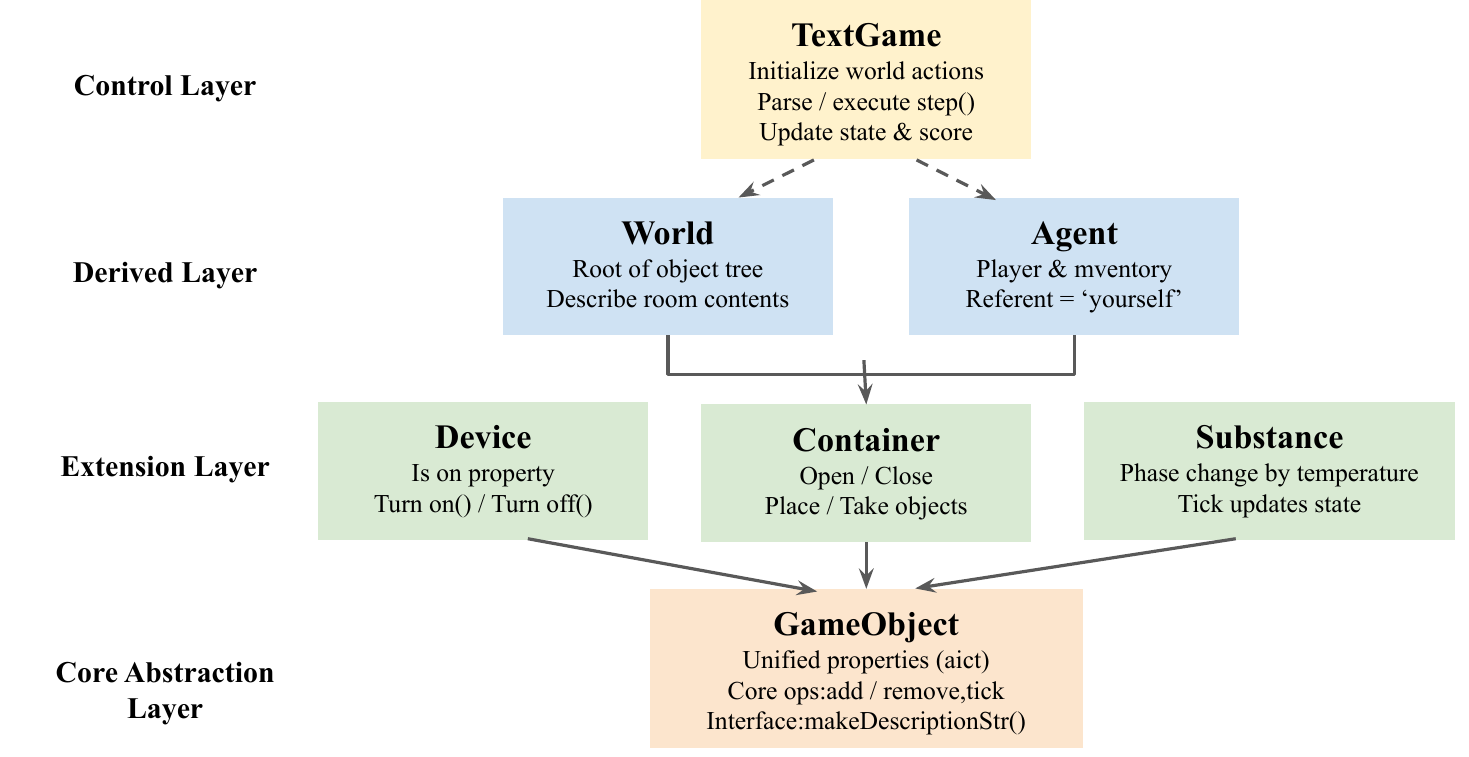}
    \caption{\small Layered architecture of the GameBasic library. Lower layers provide unified abstractions and reusable modules, while the Derived Layer defines the world and the agent. At the top, TextGame orchestrates these entities to execute the game. Arrows denote inheritance or dependency, and specifically a control relation from TextGame to World and Agent.}
    \label{fig:GameBasic}
\end{figure*}
Similar to the original \href{https://github.com/cognitiveailab/BYTESIZED32}{BYTESIZED32}, the \textbf{RefactoredByteSized32} also contains 32 common-sense task-specific text games like\textit{ washing dishes}, \textit{boiling water}. Each text game is accompanied by a \textit{\textbf{task specification}}, which provides a structured summary of the interactive environment and game objectives, including:
\begin{itemize}[nosep,leftmargin=*]
  \item \textbf{Task Description:} The natural language description of the task goal the agent must achieve, (e.g., \textit{washing dirty dishes using a dishwasher}).
  \item \textbf{Task-Critical Objects:} The set of objects indispensable for solving the task (e.g., \textit{dishes, dish soap, dishwasher}).
  \item \textbf{Actions:}The valid operations available to the agent, such as \textit{opening/closing} containers.
  \item \textbf{Distractors:} Objects or actions that are irrelevant to the task or deliberately increase its difficulty (adding \textit{food} that creates additional dirty dishes).
  \item \textbf{Solution:}A high-level procedural outline describing a canonical way to complete the task, such as opening the dishwasher.
\end{itemize}
\subsection{Refactored Code Structure}
\label{sec:gamebasic}
Each task specification of the text game encodes a goal-oriented task that requires agents to manipulate objects, containers, and devices through naturalistic action sequences shown in \autoref{fig:examplegame} and \autoref{tab:comparegame}. It is worth noting that we show in \autoref{context limits} the number of examples that can fit in the context window of various LLMs. Compared with Bytesized32, the \textbf{Bytesized32Refactored} corpus allows substantially more examples to be packed into the prompt as demonstrations under different context limits.

\begin{table*}[htbp]
\centering
\small
\caption{Comparison of BYTESIZED32 and Refactored Game Statistics (per game)}
\label{tab:comparegame}
\renewcommand{\arraystretch}{1}
\resizebox{0.8\linewidth}{!}{ 
\begin{tabular}{lccc}
\toprule
\textbf{Statistic} & \textbf{BYTESIZED32} & \textbf{BYTESIZED32Refactored} & \textbf{$\Delta$} \\
\midrule
\textbf{Lines of Python code}   & 618.1 & 303.19 & $\textbf{-50.9}$ \\
Lines of comments      & 198.1 & 198.1 &  $0$ \\
\textbf{Tokens per Game} & 6792  & 2896 & $\textbf{-3896}$ \\
Action verbs           & 9.8   & 9.8 & $0$ \\
Valid actions          & 306.6 & 306.6 & $0$  \\
Object classes         & 5.4   & 5.6 & $+0.2$ \\
Object instances       & 7.4   & 6.5 & $-0.9$ \\
Expert path length     & 12.8  & 12.8 & $0$ \\
\bottomrule
\end{tabular}}
\end{table*}
\subsubsection{GameBasic.py Library}
\label{sec: GameBaisc}
In the original ByteSized32 implementation, each game file embedded the complete implementation of multiple base classes such as \textit{GameObject}, \textit{Container}, \textit{World}, \textit{Agent}, and \textit{TextGame}. While this approach ensures code independence, it results in significant maintenance costs. We developed \verb|GameBasic.py| in \textbf{RefacoredByteSized32}, a centralized abstraction library that encapsulates common components into a modular code structure shown in Figure~\ref{fig:GameBasic}. 
Instead of these original classes in each game file, the \verb|GameBasic.py| provides a unified interface and shared logic for core functionalities, which defines seven base class abstractions: \textit{GameObject} (root class for all entities), \textit{Container} and \textit{Device} (interactive objects), \textit{Substance} (physical state modeling), \textit{World} and \textit{Agent} (environment and player), and \textit{TextGame} (template for task execution) by inheriting from these base classes. Compared to the code structure of the original ByteSized32, the \verb|GameBasic.py| brings two core contributions to the RefacoredByteSized32:
\begin{itemize}[nosep,leftmargin=*]
    \item \textbf{Consistency:} All games inherit from a unified parent class interface, ensuring structural uniformity. Developers can implement task-specific logic by simply extending these base classes, significantly reducing development complexity. These shared components in \verb|GameBasic.py| are centralized and maintained in a single location, drastically reducing repetitive code and improving maintainability.
    \item \textbf{Flexibility and Scalability:}
    Not only provides generic class implementations but also reserves function interfaces (e.g., \texttt{initializeWorld()}, \texttt{getTaskDescription()}). This design enables each game to implement customized logic with minimal development overhead, thereby enhancing extensibility and adaptability for future tasks.
\end{itemize}
\subsubsection{Code Modularity Optimization}
\label{sec: eachgamemodular}
In addition to the development of \verb|GameBasic.py|, our refactoring emphasized structured logic and loop mechanisms, leveraging higher-level function encapsulation and abstraction to replace repetitive implementations, underwent further code optimization like object initialization and action handling:
\begin{itemize}[nosep,leftmargin=*]
    \item \textbf{Action Parsing with \texttt{action\_map}:} In the original ByteSized32, the \texttt{step()} function relied on lengthy \texttt{if/elif} chains to dispatch actions. After the refactored code replaces this with an \texttt{action\_map} dictionary, significantly improving extensibility and simplicity by mapping actions directly to their corresponding handlers.
    \item \textbf{Efficient String Construction:} 
    In the original ByteSized32, \texttt{makeDescriptionStr()} relied on verbose string concatenation, whereas the refactored version uses \texttt{join()} and conditional expressions for cleaner and more efficient code.
\end{itemize}

\begin{table*}[]
\centering
\small
\caption{Number of examples that can fit in the context window of various LLMs. Compared with Bytesized32, Bytesized32Refractored can fit more examples into the prompt as demonstrations under various context limits.}
\label{context limits}
\renewcommand{\arraystretch}{1.3}
\resizebox{0.6\linewidth}{!}{ 
\begin{tabular}{lccc}\toprule
                      & GPT-4o & Qwen3-32B & Llama3.2-1b-Q \\\midrule
Context Length        & 128k   & 32k  & 8k \\
Bytesized32           & 18     & 4   & 1    \\
Bytesized32Refactored & \textbf{44} & \textbf{9} & \textbf{2}\\
\bottomrule
\end{tabular}}
\end{table*}

\section{Experiment}
\label{sec: results}
\subsection{Experiment Setup}
We demonstrate the utility of the \textbf{ByteSized32Refactored} compared with the original ByteSized32 by evaluating the quality of games generated by SOTA large language model (i.e., GPT-4o) on both corpora.
Specifically, each evaluation task provides the model with 1) a one-shot example of a game in the corpus; 2) a task specification drawn from an unseen evaluation set that asks the model to generate a new game following the specification. 

We follow the original ByteSized32 to randomly select the one-shot example and adopt the same prompt for task specification. However, while each example is presented as a single code snippet in the original corpus, in \textbf{ByteSized32Refactored}, we present the one-shot example with 1) the code and corresponding descriptions of \texttt{GameBasic.py} that informs the model of the unified interface and shared logic for game construction, and 2) the actual game code of the example showcasing how to properly leverage the classes in \texttt{GameBasic.py} to generate a game.
We provide complete details of prompts and models in \autoref{Promptdesign} and \autoref{sec:modelsetting}, respectively.

\subsection{Evaluation Metrics}
To ensure fair comparison with the original ByteSized32, we follow the original ByteSized32 to assess the quality of generated games from the following four dimensions:\\
\textbf{Technical Validity} assesses whether core mechanisms of the game run error in Python, which includes environment initialization, valid action generation, and state updates. We explore a trajectory-based method that action sequences from the initial state up to depth three (max 100 actions each) to detect runtime errors and logic inconsistencies, thereby ensuring stability and systematic verification.\\  
\textbf{Specification Compliance} measures whether a generated game satisfies its task requirements by verifying the presence and correctness of required objects, actions, and other elements. Compliance is evaluated through automatic matching against the task specification, ensuring fidelity to the intended design.\\
\textbf{Physical Reality Alignment} evaluates whether game actions respect basic physical constraints (e.g., opening a container before placing objects inside). Using trajectory-based exploration up to depth three with 100 sampled trajectories, we evaluate each step through binary judgments with justifications, ensuring logical grounding in real-world principles.\\
\textbf{Winnability} focuses on whether the generated game is winnable, i.e., whether at least one complete sequence of actions can lead to successful task completion in the game.\\

\section{Results and Analysis}
We evaluate all generated games by GPT-4o on \textbf{ByteSized32Refactored} and the original Bytesized32, reporting results before and after self-reflection to investigate whether our providing refactorization meaningfully change how LLMs build world modeling.
\begin{table}[htbp]
  \centering
  \small 
  \setlength{\tabcolsep}{4pt} 
  \caption{Technical validity of GPT-4o on \textbf{ByteSized32Refactored} across reflection steps (0–3)}
  \begin{tabularx}{\linewidth}{p{3.2cm}YYYY}
    \toprule
    \makecell{\textbf{Technical Validity}\\\textbf{Measurement}} 
      & \multicolumn{4}{c}{\textbf{Number of Reflections}} \\
    \cmidrule(lr){2-5}
      & 0 & 1 & 2 & 3 \\
    \midrule
    Game Initialization & 56.25\% & 81.25\% & 85.42\% & 85.42\% \\
    Vaild Actions      & 55.21\% & 62.50\% & 69.79\% & 70.83\% \\
    Runnable Game       & 17.71\% & 39.58\% & 55.21\% & 61.46\% \\
    \bottomrule
  \end{tabularx}
  \label{tab:validity_gpt4o_new}
\end{table}

\begin{table}[htbp]
  \centering
  \small 
  \setlength{\tabcolsep}{4pt} 
  \caption{Technical validity of GPT-4o on \textbf{Bytesized32} across reflection steps (0–3)}
  \begin{tabularx}{\linewidth}{p{3.2cm}YYYY}
    \toprule
    \makecell{\textbf{Technical Validity}\\\textbf{Measurement}} 
      & \multicolumn{4}{c}{\textbf{Number of Reflections}} \\
    \cmidrule(lr){2-5}
      & 0 & 1 & 2 & 3 \\
    \midrule
    Game Initialization & 91.67\% & 94.79\% & 95.83\% & 95.83\% \\
    Vaild Actions       & 75.00\% & 90.62\% & 90.62\% & 90.62\% \\
    Runnable Game       & 48.96\% & 75.00\% & 80.21\% & 82.29\% \\
    \bottomrule
  \end{tabularx}
  \label{tab:validity_gpt4o_old}
\end{table}
\paragraph{Technical Validity} The results of the technical validity are summarized in \autoref{tab:validity_gpt4o_new} and \autoref{tab:validity_gpt4o_old}. On all of \textit{game initialization}, \textit{valid actions} and \textit{runnable game}, the generated games demonstrate increasing quality with more founds of reflection on both \textbf{Bytesized32Refactored} and the original Bytesized32, showcasing the effectiveness of reflection in improving the quality of game generation across original and refactored codebase. However, on all of the three metrics, the game quality in \textbf{Bytesized32Refactored} down-performs the counter-part in the original Bytesized32 across all rounds of reflection. The degredation of quality reveals that the refactored code presents new challenges for LLM to reasoning over the hierarchical code structure and perform more complex class-based code generation.
\begin{figure*}[htbp]
  \centering
  \begin{minipage}{0.48\textwidth}
    \centering
    \includegraphics[width=\linewidth,trim=20 10 60 40,clip]{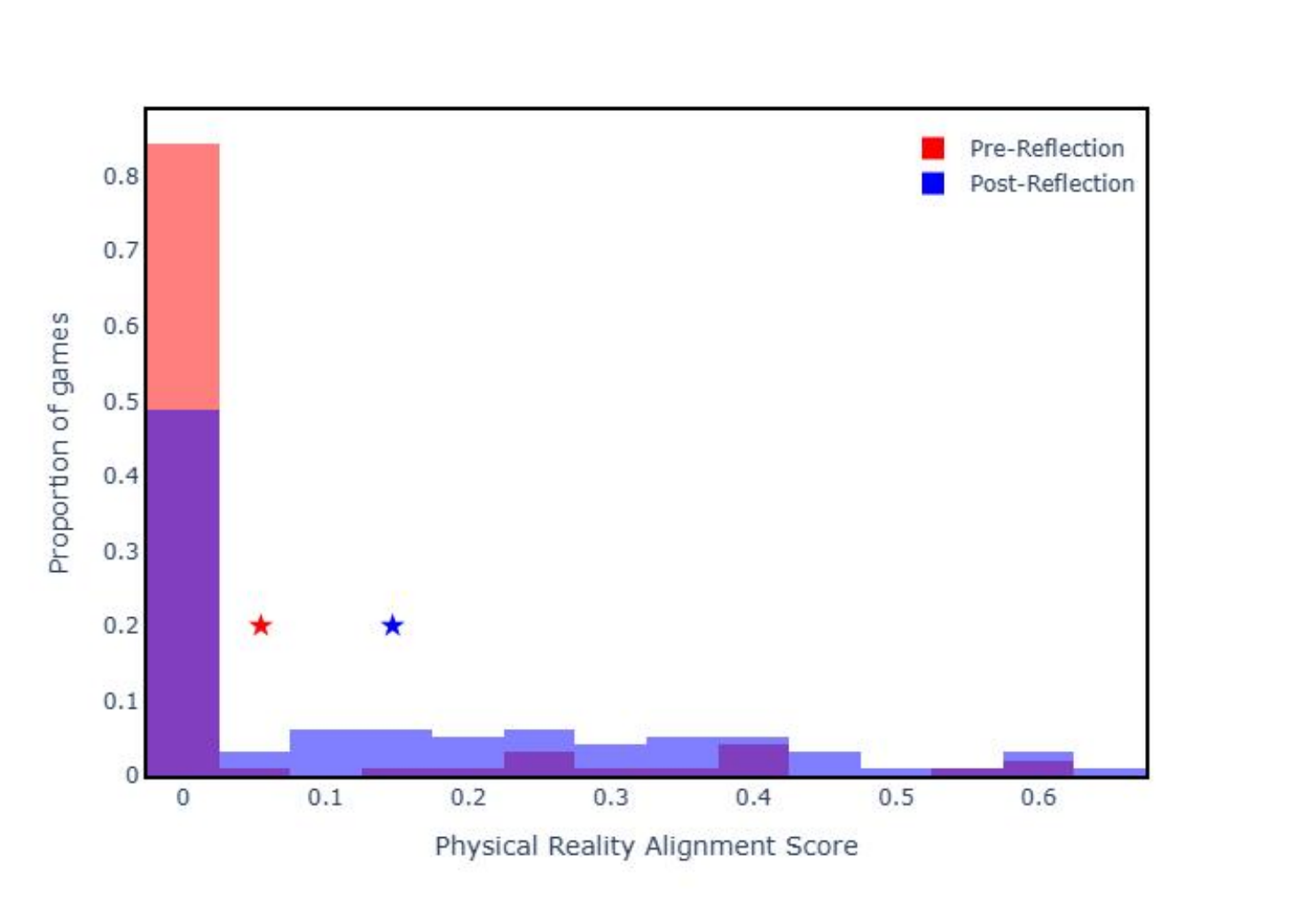}
    \caption{Histogram of physical reality alignment scores for GPT-4o on \textbf{ByteSized32Refactored}before(red) and after reflection(blue). Asterisk represent average scores(0.055 pre-reflection,0.147 post-reflection)}
    \label{fig:physicalreality_gpt4o}
  \end{minipage}
  \hfill
  \begin{minipage}{0.48\textwidth}
    \centering
    \includegraphics[width=\linewidth,trim=20 10 60 40,clip]{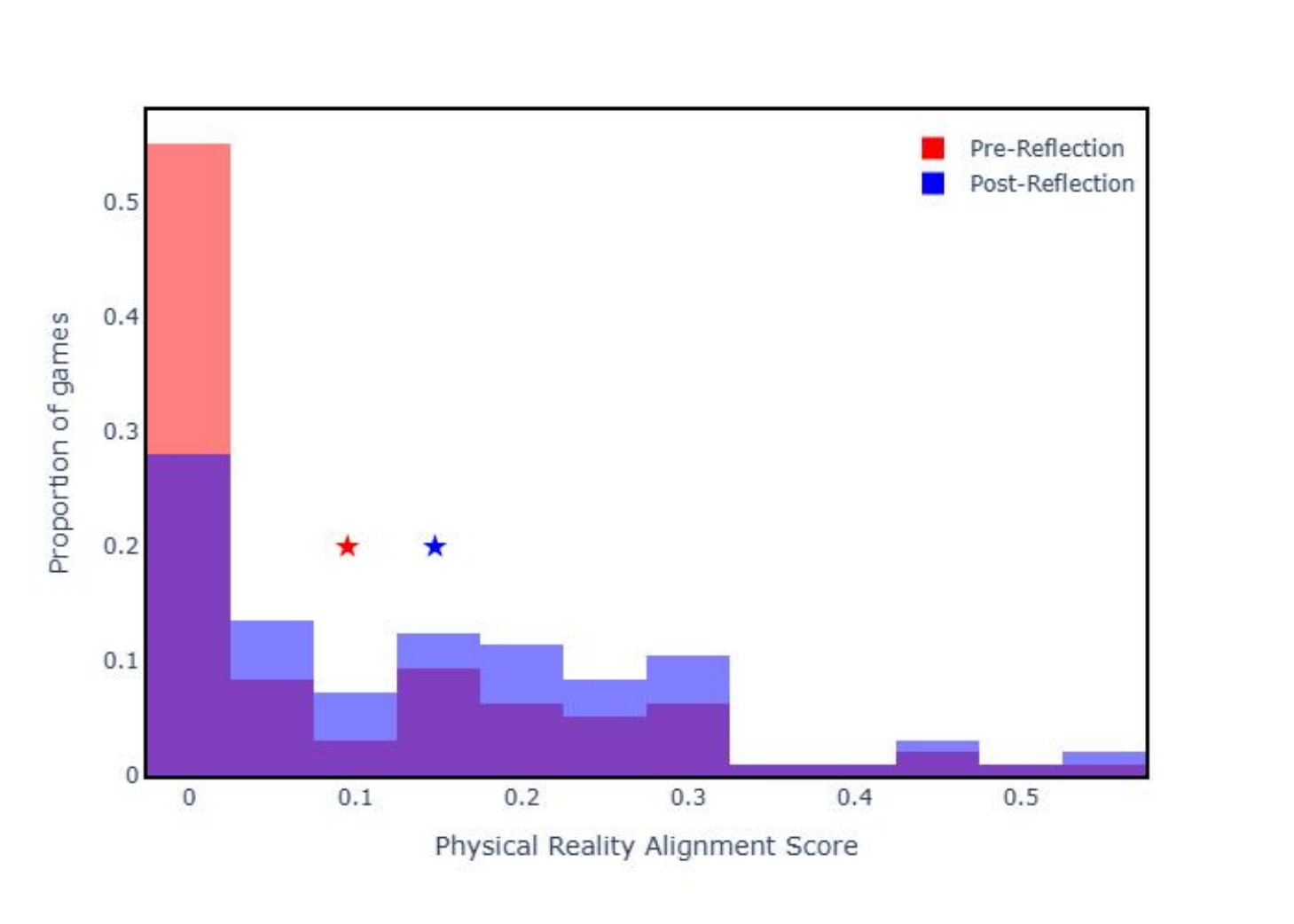}
    \caption{Histogram of physical reality alignment scores for GPT-4o on \textbf{ByteSized32} before(red) and after reflection(blue). Asterisk represent average scores(0.095 pre-reflection,0.148 post-reflection)}
    \label{fig:physicalreality_gpt4o_old}
  \end{minipage}
\end{figure*}
\paragraph{Specification Compliance and Winnability} We report the specification compliance and winnablity performance of GPT-4o under the refactored and original corpus in \autoref{tab:spec-winn_gpt4o} and \autoref{tab:spec-winn_gpt4o_old}.
\begin{table}[htbp]
  \centering
  \small
  \setlength{\tabcolsep}{4pt} 
  \caption{Specification compliance and winnability for GPT-4o on \textbf{ByteSized32Refactored} before/after reflection.}
  \label{tab:spec-winn_gpt4o}
  \begin{tabularx}{\linewidth}{L{0.48\linewidth}YY Y}
    \toprule
    \makecell{\textbf{Measurement}} &
    \multicolumn{2}{c}{\textbf{Reflection}} & \textbf{$\Delta$} \\
    \cmidrule(lr){2-3}
     & \textbf{Before} & \textbf{After} &  \\ 
    \midrule
   \itshape Specification Compliance \\
    \quad Task-critical objects & 100.0\% & 100.0\% & 0.0\%  \\
    \quad Task-critical actions & 93.75\% & 93.75\% & 0.0\%  \\
    \quad Distractors          & 31.25\% & 28.12\% & -3.13\%  \\
    \addlinespace[2pt]
    \cmidrule(lr){1-4}
    \addlinespace[2pt]
    Winnability                 & 33.3\% & 54\% & +20.7\%  \\
    \bottomrule
  \end{tabularx}
\end{table}
\begin{table}[htbp]
  \centering
  \small
  \setlength{\tabcolsep}{4pt} 
  \caption{Specification compliance and winnability for GPT-4o on \textit{\textbf{ByteSized32}} before/after reflection.}
  \label{tab:spec-winn_gpt4o_old}
  \begin{tabularx}{\linewidth}{L{0.48\linewidth}YY Y}
    \toprule
    \makecell{\textbf{Measurement}} &
    \multicolumn{2}{c}{\textbf{Reflection}} & \textbf{$\Delta$} \\
    \cmidrule(lr){2-3}
     & \textbf{Before} & \textbf{After} &  \\ 
    \midrule
   \itshape Specification Compliance \\
    \quad Task-critical objects & 100.0\% & 100.0\% & 0.0\%  \\
    \quad Task-critical actions & 90.62\% & 90.62\% & 0.0\%  \\
    \quad Distractors          & 56.25\% & 53.12\% & -3.13\%  \\
    \addlinespace[2pt]
    \cmidrule(lr){1-4}
    \addlinespace[2pt]
    Winnability                 & 20\% & 34\% & +14\%  \\
    \bottomrule
  \end{tabularx}
\end{table}
Specifically, on both corpus, the compliance of task-critical objects and actions and winnability increase with three rounds of reflection whereas the number of distractors slightly drops. Across inital generation and reflections, the generated game on \textbf{Bytesized32Refactored} consistently outperforms games on the original counterpart over object/action compliance and winnability, showcasing that GPT-4o is able to generate more reasonable games that better comply with specifications on \textbf{Bytesized32Refactored}, highlighting the effectiveness of our refactorization.
\paragraph{Physical Reality Alignment} \autoref{fig:physicalreality_gpt4o} and \autoref{fig:physicalreality_gpt4o_old} illustrate how the generated games align with physical reality on both corpora. On average, comparing with performance on the original corpus, GPT-4o struggles to output physically-aligned games on \textbf{Bytesized32Refactored} on its initial generation but can attain a similar level of alignment after reflection. By comparing the two histograms, we observe that the higher abstraction and compressed structure of Refactored ByteSized32 initially make it more challenging for GPT-4o to capture physical consistency. This is evident in the distribution, which initially clusters near zero with few mid- or high-scoring samples. However, reflection significantly reduces low scores and gradually shifts the distribution toward the mid- and high-range. The results indicate that with reflection, the refactored corpus does not hinder GPT-4o from yielding physically-aligned text games compared to the original corpus.

\paragraph{Summary of results} In summary, our findings across the four evaluation dimensions highlight two key insights in the evaluation of LLMs for text-game generation: (1) The abstracted interfaces in \verb|GameBasic.py| pose a greater challenge for GPT-4o than the redundant implementation in the original corpus, which embeds more explicit action cues that enables models to pass validity checks without deep reasoning. By contrast, abstraction increases reasoning demands and raises the generation barrier, making GPT-4o more prone to omitting necessary components or producing invalid structures during initial generation. (2) On the other hand, our hierarchical abstractions enables LLMs to generate valid text games that 1) are much more complied with given specification; 2) possess high-quality underlying game logic to ensure winnability; and 3) do not sacrifice the alignment with physical reality under reflection.

\section{Conclusion}
In this work, we presented \textbf{ByteSized32Refactored}, a modular and extensible reimplementation of the original ByteSized32 corpus, designed to advance the exploration of text game generation with large language models (LLMs). \verb|GameBasic.py| not only reduces code redundancy in text-based games, but also decouples basic components, such as gameobject, container, and agent, from specific task logic, providing a unified framework for future text-based games. This design makes common text-based game features easy to extend and opens up the possibility of unlimited future expansion. With our refactored design, \textbf{ByteSized32Refactored} can be easily extended to new game scenarios by reusing common game logic and substantially increasing the number of examples in a limited LLM context window. Our experiments with GPT-4o demonstrate mixed performance of the generated games, revealing both the effectiveness of the refactored codebase and new challenges presented by the hierarchical design. Overall, our findings emphasize both the promise and limitations of modular design and foundational libraries in enabling LLMs to handle complex environment specifications, paving the way for future progress at the intersection of world modeling and code generation.
\section*{Limitations}
Our work has several limitations that should be acknowledged.
\begin{enumerate}[nosep, leftmargin=*]
    \item The \textbf{ByteSized32Refactored} corpus alters token distribution and code structure. This raises the possibility that observed improvements stem from structural biases rather than genuine advancements in modeling ability.
    \item Our evaluation metrics—validity, compliance, alignment, and winnability—serve as approximations of world modeling but have limited construct validity, being susceptible to framework biases and coverage gaps. 
    \item While reflection enhances measurability, it does not directly improve ability. Future work should aim to strengthen validity through invariant testing and randomization, disentangle compliance dimensions, clarify alignment with multi-judge evaluations, and expand winnability metrics with broader coverage and fuzzing techniques.
    \item Finally, we perform reflection by regenerating the complete program at each step and target only a single error at a time. This process could be made more efficient by outputting only a code diff and batching multiple errors at once.
\end{enumerate}

\bibliography{reference}

\appendix

\label{sec:appendix}

\section{Model Setting}
\label{sec:modelsetting}
In this work, we make extensive use of OpenAI's API. In all our experiments, we keep the following hyperparameters constant:In this work, we make extensive use of OpenAI's API. In all our experiments, we keep the following hyperparameters constant:

\subsection{GPT-4o Model setting}
\begin{itemize}[nosep, leftmargin=*]
    \item temperature=0.0
    \item top-p=1
    \item frequency-penalty=0.0
    \item persence-penalty=0.0
\end{itemize}
\subsection{GPT-5 Model setting}
\label{sec:gpt5modelsetting}
\begin{itemize}[nosep, leftmargin=*]
    \item temperature=1
    \item top-p=1
    \item frequency-penalty=0.0
    \item persence-penalty=0.0
\end{itemize}
\section{An Example of Playthrough in Bytesized32Refactored Corpus}

\begin{figure}[htbp]
    \centering
    \includegraphics[width=\linewidth]{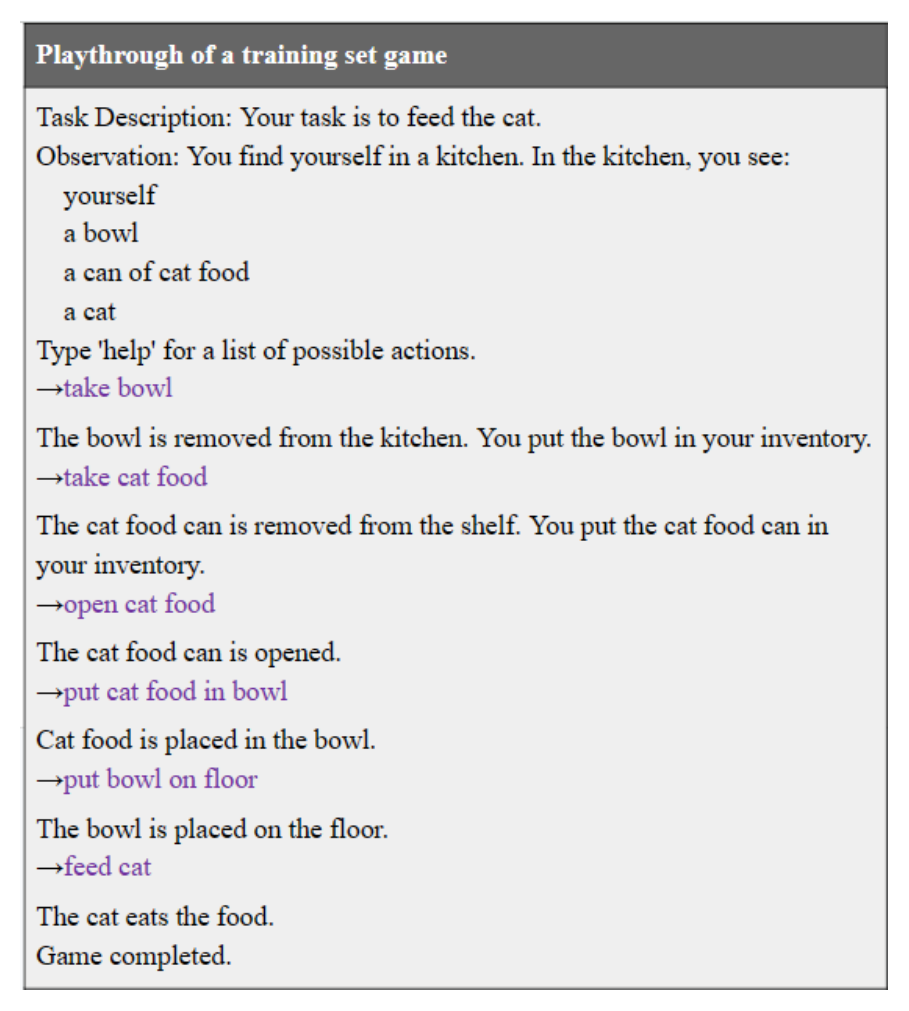}
    \caption{An example of Playthrough.}
    \label{fig:examplegame}
\end{figure}
\newpage
\section{GPT-5 Experiment error and analysis}

\begin{figure}[htbp]
    \centering
    \includegraphics[width=0.8\linewidth]{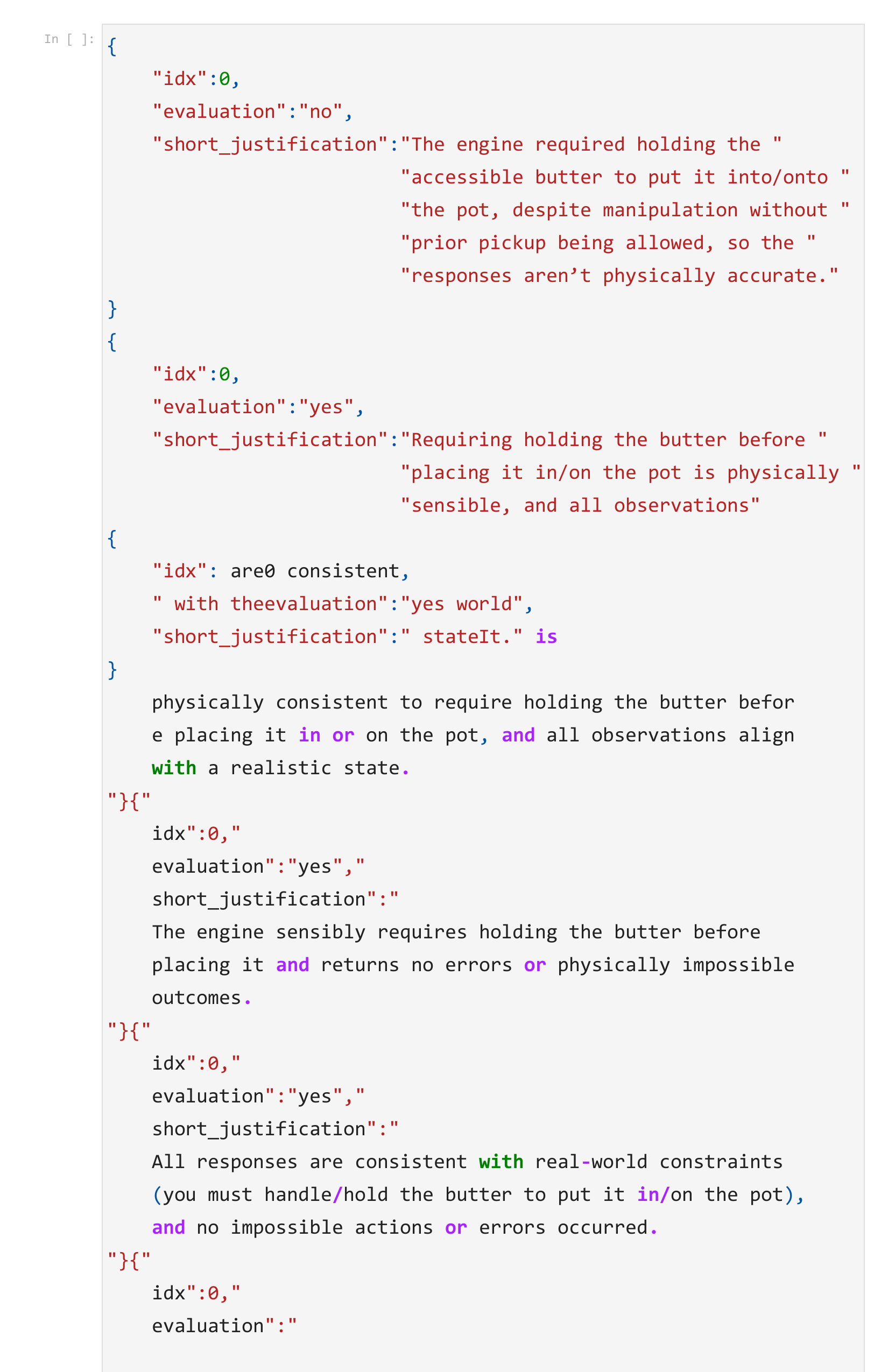}
    \caption{GPT-5-Alignment-response-bug.}
    \label{fig:bug}
\end{figure}
\begin{figure}[htbp]
    \centering
    \includegraphics[width=0.8\linewidth,trim=0cm 32cm 0cm 0cm,clip]{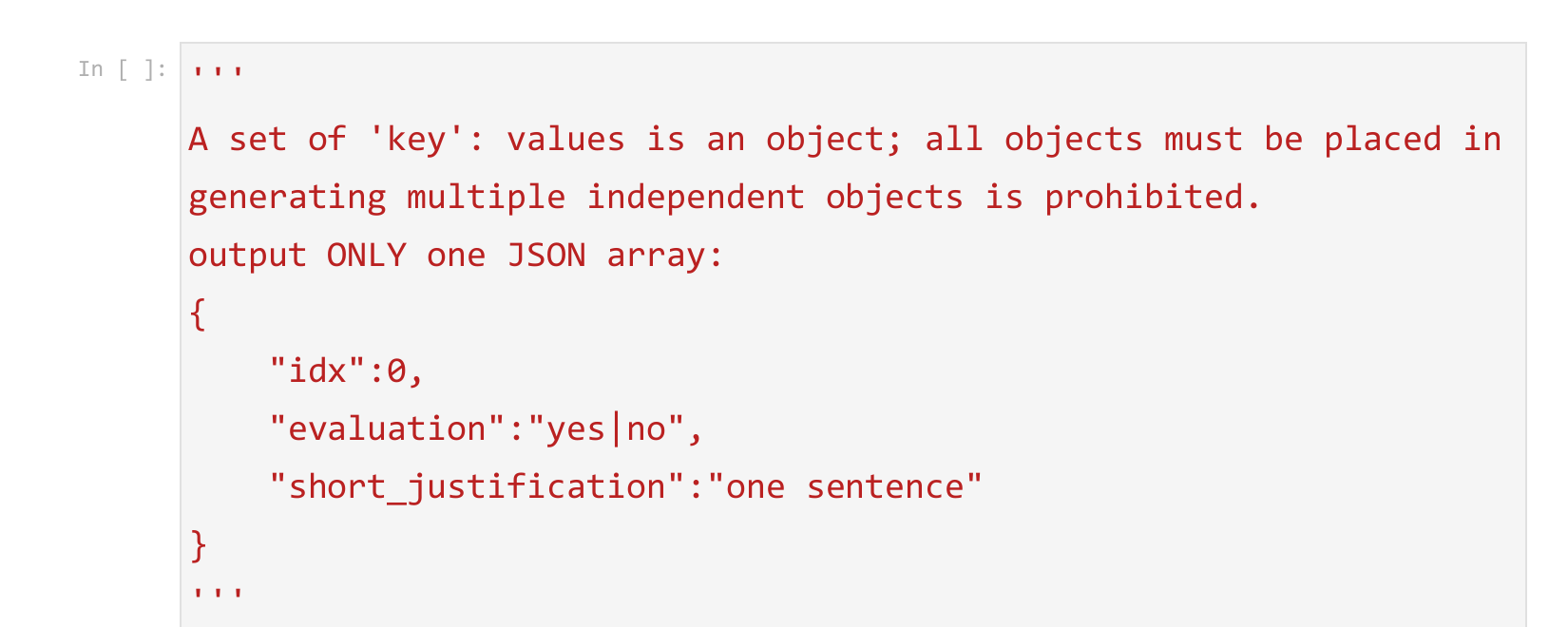}
    \caption{GPT5Alignmentpromptincrease.}
    \label{fig:addprompt}\label{fig:addprompt}
\end{figure}
\section{GPT-5 Experiment Results and Discussion}
Similarly, we keep \textbf{GPT-5} constant to evaluate all generated games in ByteSized32Refactored and the original ByteSized32. While we were able to obtain valid game generations from GPT-5, we encountered challenges during the evaluation and reflection process, as GPT-5 was unable to execute the provided evaluation and reflection code and prompt successfully. It is worth that rather than indicating a weakness of the RefactoredByteSized32 corpus, the incompatibility reflects GPT-5’s different API constraints (e.g., output formatting, limited batch generation, timeout issues) and underscores the need for more robust, model-agnostic evaluation pipelines, the failure and experiment process for  GPT-5 which highlights the practical limitations of our current evaluation methods when applied to LLMs and to supr futher development at the juncture of world modeling and code generation.\\ 
\textbf{Challenge 1:Output Format} GPT-5 often deviated from the expected output format required by the evaluation scripts, leading to parsing errors and failed executions during the \textit{Physical Reality Alignment} evaluation metrics. Despite prompt adjustments shown in new add Prompt~\ref{fig:addprompt}, GPT-5 failed to produce outputs in the expected fixed JSON format with syntax errors and logical inconsistencies resulting in Figure~\ref{fig:bug}. The error include \textbf{\textit{JSON Structure Errors}}, such as missing or mismatched brackets (e.g., a “{” without a corresponding “}”), rendering the output unparsable; \textbf{\textit{Syntax Irregularities}}, such as concatenated words (e.g., "theevaluation") without necessary spaces, causing readability and parsing issues; and \textbf{\textit{Structural Disarray}}, where outputs exhibited syntactic and semantic incoherence, with poor sentence transitions, broken logical relationships, and contradictory semantics, resulting in outputs that lacked readability and consistency.\\
\textbf{Challenge 2: Timeouts and API  Maximum Limitations.} 
While we successfully used GPT-5 to generate all games, the reflection stage proved problematic. 
For several games (e.g., [specific games]), the multi-iteration reflection process exceeded the preset 30-minute timeout, and GPT-5’s higher latency further disrupted the iterative improvement cycle. 
To proceed with later stages, these games had to be removed from the directory to continue evaluation and reflection. 
In addition, GPT-5’s API enforced a strict maximum call limit of \(n \leq 8\), which prevented us from executing the compliance-majority-vote procedure that requires 31 generations.
\newpage
\section{Discussion-LLMs Difference: Do Newer LLMs Surpass GPT-4 in World Modeling?}
In addition to examining the impact of code structure through our Refactored vs. Original comparison, we also investigated a complementary question: do newer LLMs surpass GPT-4 in world modeling when evaluated on the same original ByteSized32 corpus? While our main focus is on the structural effects introduced by ByteSized32Refactored, this second line of analysis provides insight into model-level differences independent of corpus design. For clarity, we present the results of this comparison in a separate subsection (and provide extended details in the Appendix).
\subsection{GPT-4o vs GPT-4 on original Bytesized32}
We keep the original Bytesized32 in constant and evaluate all generated games (N = 96) in GPT-4o and GPT-4, reporting results both before and after self-reflection to investigate whether the newer GPT-4o surpasses GPT-4 in building world models without Bytesized32Refactored.\\
The results of the \textit{technical validity} are summarized in Table~\ref{tab:validity_gpt4o_old} and Table~\ref{tab:validity_gpt4_32K}.

\begin{figure}[htbp]
    \centering
    \includegraphics[width=\linewidth,trim=20 10 75 55,clip]{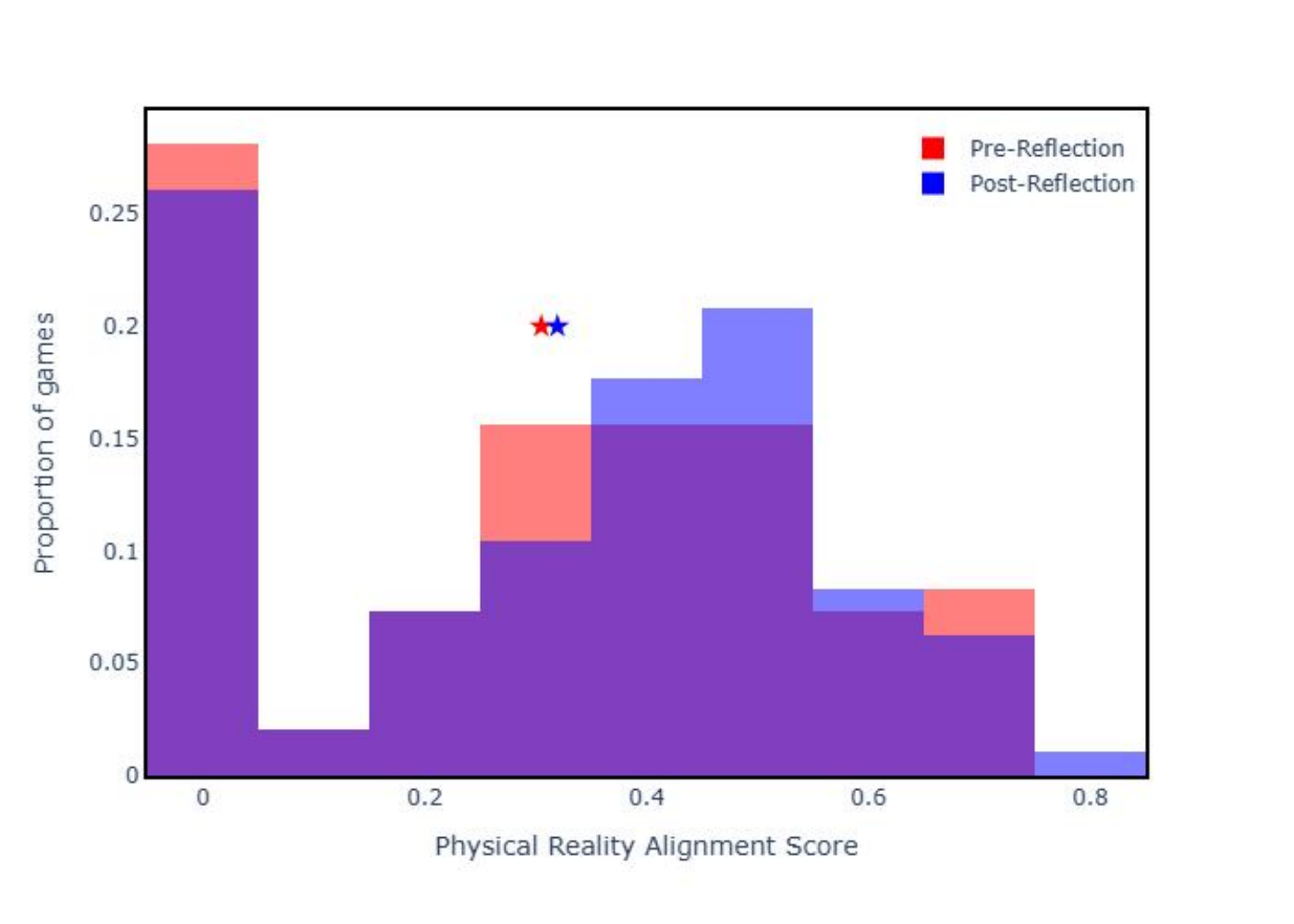}
    \caption{Histogram of physical reality alignment scores before and after reflection.}
    \label{fig:physicalreality_gpt4_32K}
\end{figure}

\begin{table}[htbp]
  \centering
  \small 
  \setlength{\tabcolsep}{4pt} 
  \caption{Technical validity of GPT-4-32k on \textbf{ByteSized32Refactored} across reflection steps (0–3).}
   \label{tab:validity_gpt4_32K}
  \begin{tabularx}{\linewidth}{p{3.2cm}YYYY}
    \toprule
    \makecell{\textbf{Technical Validity}\\\textbf{Measurement}} 
      & \multicolumn{4}{c}{\textbf{Number of Reflections}} \\
    \cmidrule(lr){2-5}
      & 0 & 1 & 2 & 3 \\
    \midrule
    Game Initialization & 85.4\% & 85.4\% & 89.6\% & 88.5\% \\
    Valid Actions       & 80.2\% & 83.3\% & 87.5\% & 88.5\% \\
    Runnable Game       & 28.1\% & 42.7\% & 51.0\% & 57.3\% \\
    \bottomrule
  \end{tabularx}
\end{table}
On technical validity, GPT-4o consistently outperforms GPT-4 across all three metrics on the original ByteSized32 corpus. On game initialization, GPT-4o achieves 91.67\% valid implementations before reflection, increasing to 95.83\% after three reflections. In contrast, GPT-4 starts lower at 85.4\% and rises to 89.6\%, showing a +4.2\% improvement but still trailing GPT-4o by a significant margin of 4.16\%. On valid actions, GPT-4o begins at 75.00\% and increases to 90.62\%, while GPT-4 starts at a lower 80.2\% and improves to 88.5\%, resulting in a substantial gap of 8.3\% after reflection. Finally, on runnable games, GPT-4o starts at 48.96\% and rises to 82.29\%, whereas GPT-4 begins at a much lower 28.1\% and only reaches 57.3\%, leaving a gap of 29.2\% after reflection.\\
Similarly to the results of the \textit{physical reality alignment},\textit{specification compliance and winnability of games} are summarized in Table~\ref{tab:spec-winn_gpt4o_old}, Table~\ref{tab:spec-winn_gpt4_32K}, Figure~\ref{fig:physicalreality_gpt4o_old} and Figure~\ref{fig:physicalreality_gpt4_32K}. 
GPT-4o improves from 9.5\% to 14.8\% (+5.3\%), while GPT-4 rises only from 30.25\% to 31.97\% (+1.72\%). Score distributions show GPT-4o produces more mid- to high-range results, whereas GPT-4 clusters at lower scores. This indicates GPT-4o benefits more from reflection and achieves a more balanced alignment.\\
\begin{table}[htbp]
  \centering
  \small
  \setlength{\tabcolsep}{4pt} 
  \caption{Specification compliance and winnability for GPT-4-32k on \textit{\textbf{Original ByteSized32}} before/after reflection.}
  \label{tab:spec-winn_gpt4_32K}
  \begin{tabularx}{\linewidth}{L{0.48\linewidth}YY Y}
    \toprule
    \makecell{\textbf{Measurement}} &
    \multicolumn{2}{c}{\textbf{Reflection}} & \textbf{$\Delta$} \\
    \cmidrule(lr){2-3}
     & \textbf{Before} & \textbf{After} &  \\ 
    \midrule
   \itshape Specification Compliance \\\itshape Specification Compliance \\
    \quad Task-critical objects & 100.0\% & 100.0\% & 0.0\%  \\
    \quad Task-critical actions & 93.8\% & 93.8\% & 0.0\%  \\
    \quad Distractors          & 21.9\% & 18.8\% & -3.1\%  \\
    \addlinespace[2pt]
    \cmidrule(lr){1-4}
    \addlinespace[2pt]
    Winnability                 & 30.2\% & 37.5\% & +7.3\%  \\
    \bottomrule
  \end{tabularx}
\end{table}
These results indicate that GPT-4o has a stronger initial capability and benefits more from self-reflection compared to GPT-4. GPT-4-32K achieves a higher initial alignment score (0.3025 vs. 0.095) but gains little from reflection (+0.0172), whereas GPT-4o starts much lower but improves more substantially (+0.053). This indicates that GPT-4-32K’s generations are already physically grounded, yet its reasoning benefits less from iterative correction, while GPT-4o, despite its weaker initial grounding, shows stronger reflection-driven reasoning. 

\section{RefactoredBYTESIZED32 PYTHON TEMPLATE in GPT-4o}
\begin{figure}[htbp]
    \centering
    \includegraphics[width=\linewidth]{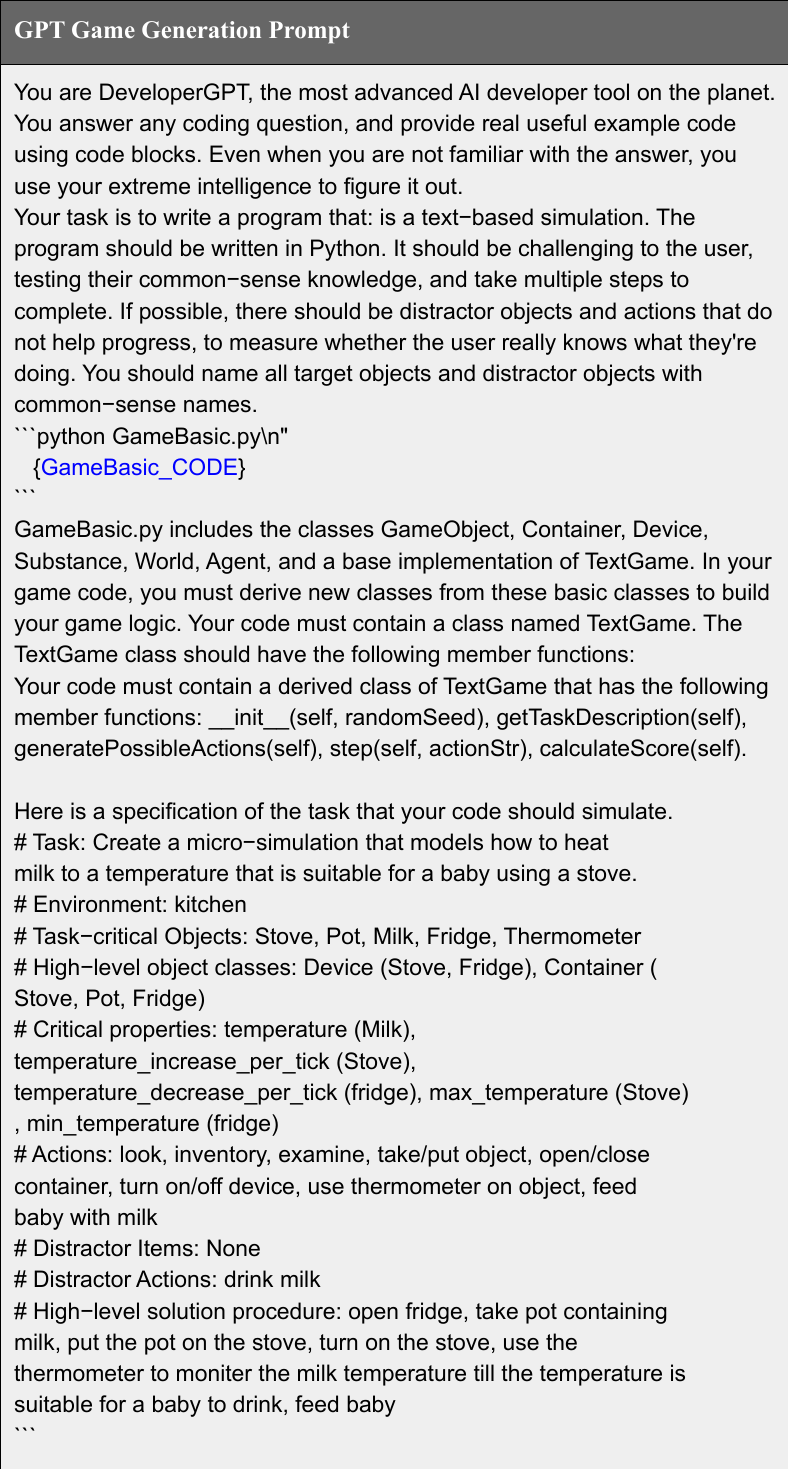}
\end{figure}
\begin{figure}[htbp]
    \centering
    \includegraphics[width=\linewidth]{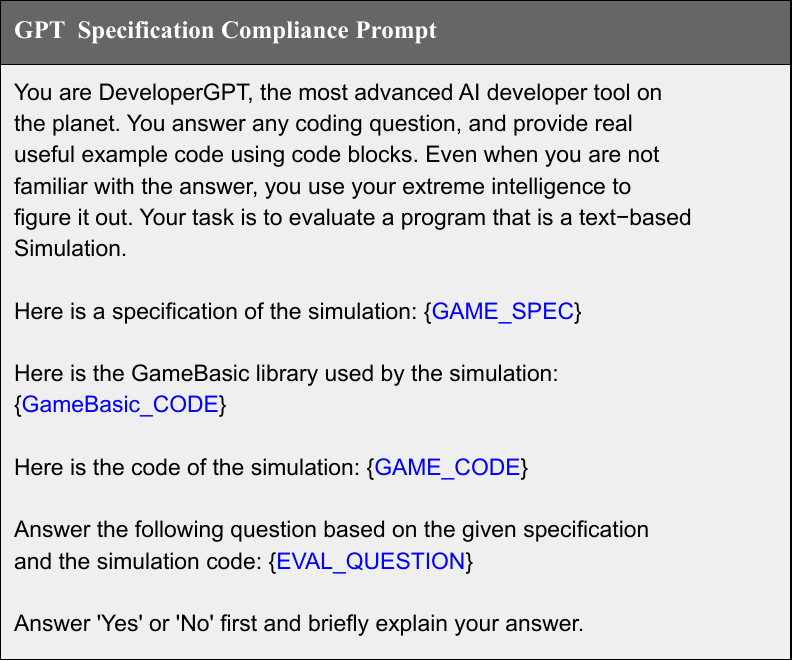}
\end{figure}
\begin{figure}[htbp]
    \centering
    \includegraphics[width=\linewidth]{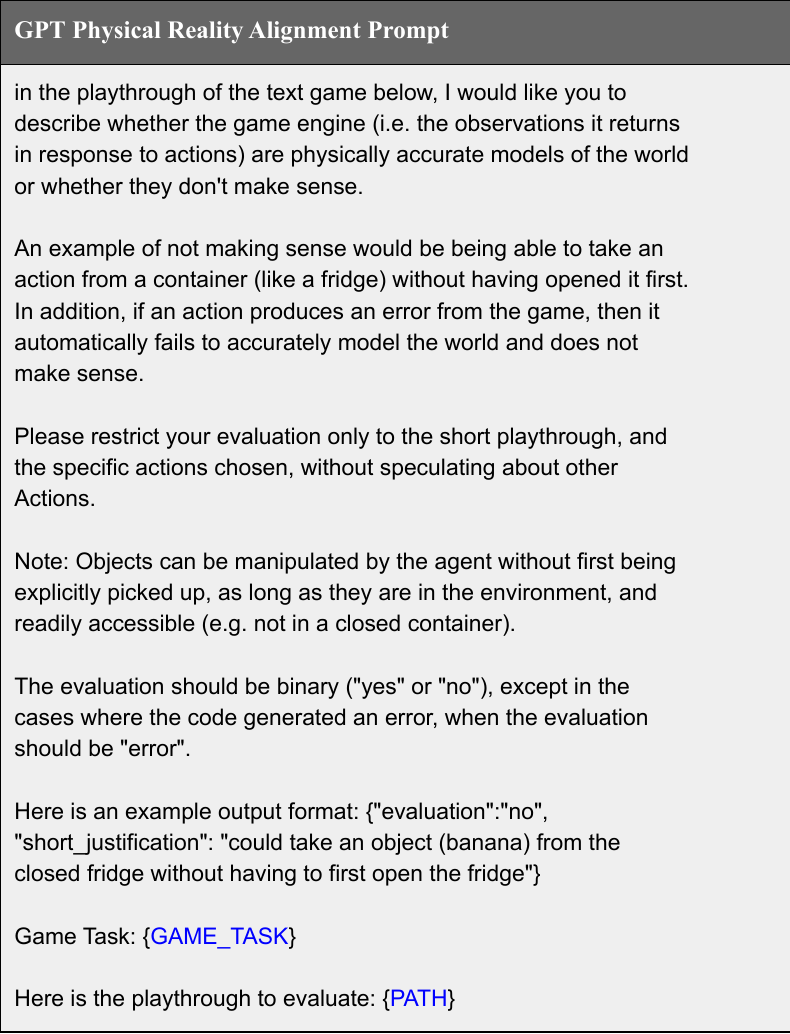}
\end{figure}
\begin{figure}[htbp]
    \centering
    \includegraphics[width=\linewidth]{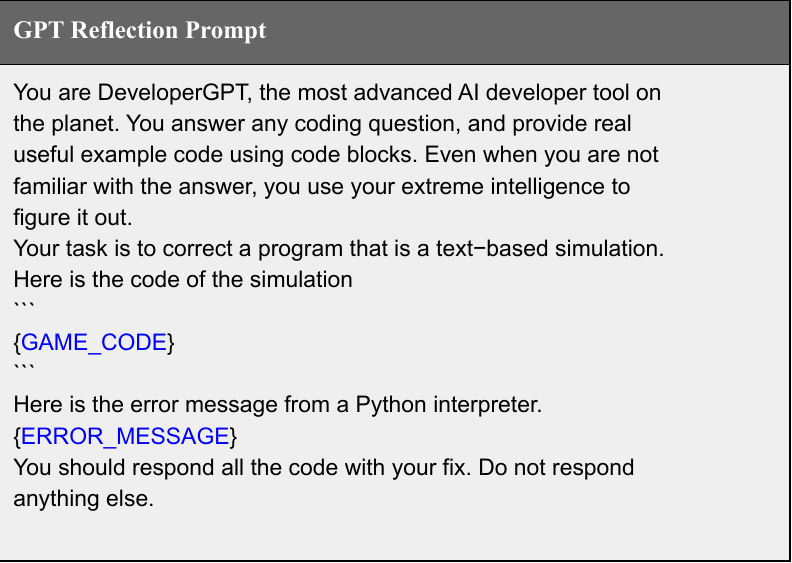}
\end{figure}
\begin{figure}[htbp]
    \centering
    \includegraphics[width=\linewidth]{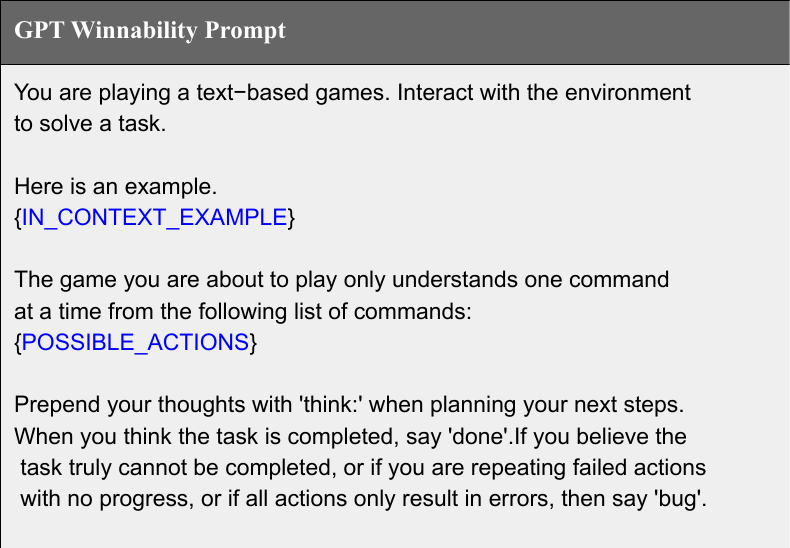}
\end{figure}
\begin{figure}[htbp]
    \centering
    \includegraphics[width=\linewidth]{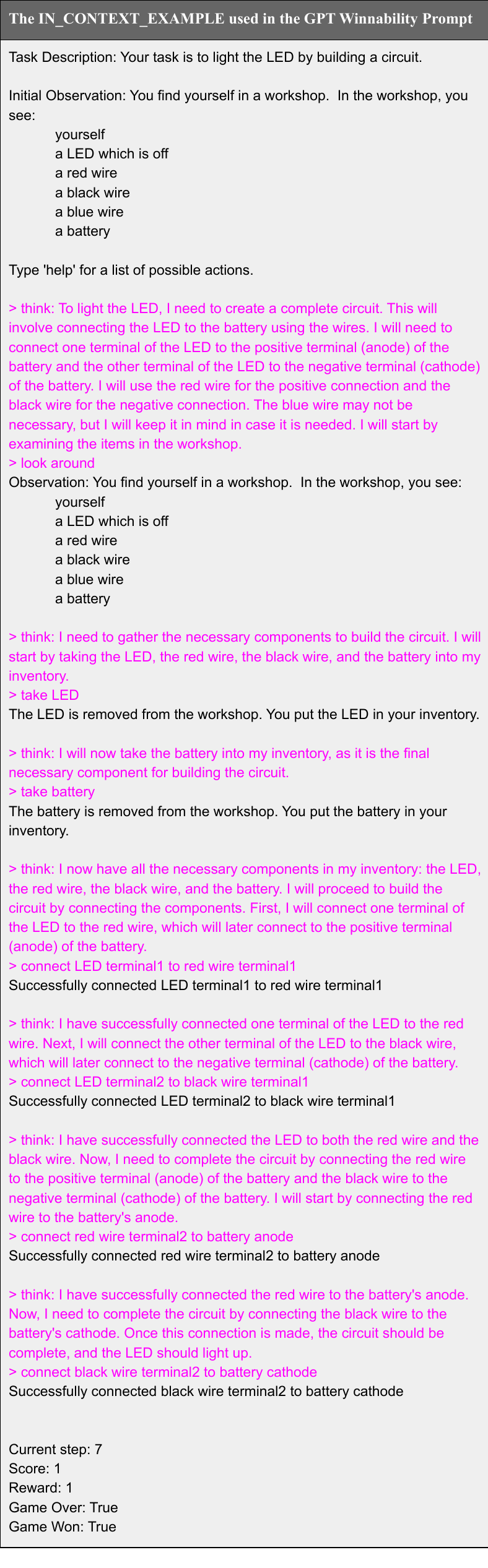}
\end{figure}

\label{Promptdesign}

\end{document}